\pdfoutput=1

\documentclass[11pt]{article}

\usepackage{acl}

\usepackage{times}
\usepackage{latexsym}

\usepackage{multirow}
\usepackage{amsmath}
\usepackage{graphicx}

\usepackage[T1]{fontenc}

\usepackage[utf8]{inputenc}
\usepackage{url}
\usepackage{microtype}

\title{An Efficient Coarse-to-Fine Facet-Aware Unsupervised Summarization Framework based on Semantic Blocks}

\author{Xinnian Liang\textsuperscript{1}\footnotemark[1], Jing Li\textsuperscript{2}, Shuangzhi Wu\textsuperscript{3}, Jiali Zeng\textsuperscript{3}, Yufan Jiang\textsuperscript{3}, Mu Li\textsuperscript{3}, Zhoujun Li\textsuperscript{1}\footnotemark[2] \\
\textsuperscript{1}State Key Lab of Software Development Environment, Beihang University, Beijing, China \\ 
\textsuperscript{2} School of Information Renmin University of China, Beijing, China \\
\textsuperscript{3}Tencent Cloud Xiaowei, Beijing, China\\ 
\texttt{\{xnliang,lizj\}@buaa.edu.cn};
\texttt{heylijing@126.com}\\
\texttt{\{frostwu,lemonzeng,garyyfjiang\}@tencent.com,limugx@qq.com};\\ } 

\begin{document}
\maketitle
\renewcommand{\thefootnote}{\fnsymbol{footnote}} 
\footnotetext[1]{Contribution during internship at Tencent Inc.} \footnotetext[2]{Corresponding Author} 
\renewcommand{\thefootnote}{\arabic{footnote}} 

\begin{abstract}
Unsupervised summarization methods have achieved remarkable results by incorporating representations from pre-trained language models. However, existing methods fail to consider efficiency and effectiveness at the same time when the input document is extremely long. To tackle this problem, in this paper, we proposed an efficient Coarse-to-Fine Facet-Aware Ranking (C2F-FAR) framework for unsupervised long document summarization, which is based on the semantic block. The semantic block refers to continuous sentences in the document that describe the same facet. Specifically, we address this problem by converting the one-step ranking method into the hierarchical multi-granularity two-stage ranking. In the coarse-level stage, we propose a new segment algorithm to split the document into facet-aware semantic blocks and then filter insignificant blocks. In the fine-level stage, we select salient sentences in each block and then extract the final summary from selected sentences. We evaluate our framework on four long document summarization datasets: Gov-Report, BillSum, arXiv, and PubMed. Our C2F-FAR can achieve new state-of-the-art unsupervised summarization results on Gov-Report and BillSum. In addition, our method speeds up 4-28 times more than previous methods.\footnote{\url{https://github.com/xnliang98/c2f-far}}
\end{abstract}

\section{Introduction}
The text summarization task aims to condense a document or a set of documents into several sentences and keep the primary information. 
Recent years, both supervised \cite{liu-lapata-2019-text,liu-liu-2021-simcls,liu-etal-2021-refsum} and unsupervised \cite{zheng-lapata-2019-sentence,dong-etal-2021-discourse,liang-etal-2021-improving,9664266} methods have made significant improvements over short documents with the development of semantic representations from Pre-trained Language Models (PLMs). 
Due to the noise and complexity of the increased input and output length, long-form document summarization is still a challenge \cite{tay2021long,akiyama-etal-2021-hie,grail-etal-2021-globalizing}.  Compared with supervised one, unsupervised methods do not rely on large amounts of labeled data and have no limitation on input length. 
In addition, unsupervised methods can be easily adapted to data from different domains, types, and languages. In this paper, we focus on unsupervised extractive methods for long document summarization.

\begin{figure*}[]
    \centering
    \includegraphics[width=\textwidth]{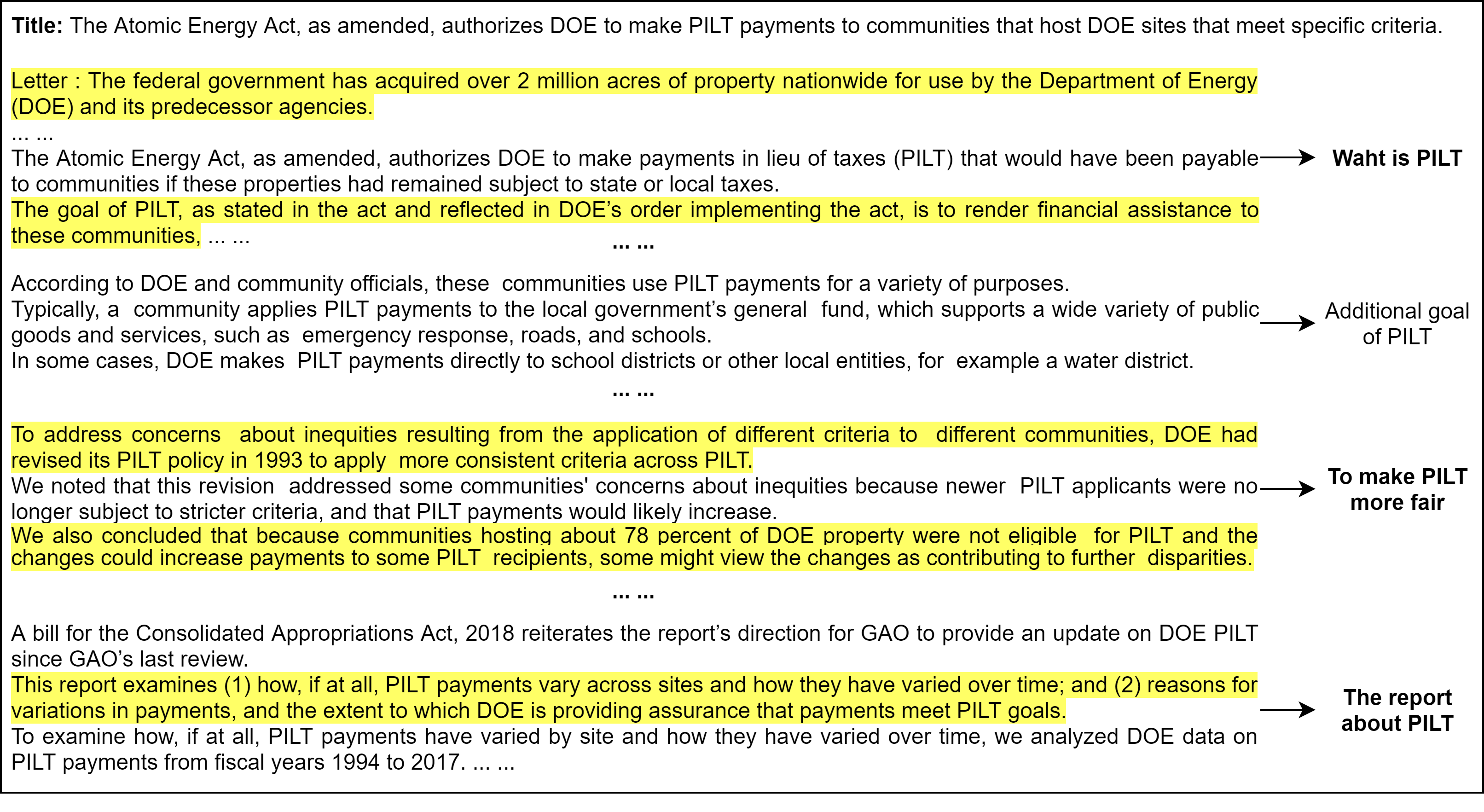}
    \caption{An example from the Gov-Report dataset to introduce the process of our method. ``...'' refers to the omissions of context sentences due to space limitations. Highlight sentences refer to the final extracted summary sentences. The content of the arrow pointed is the facet description of the left semantic block. Bold facets represent vital facet-aware semantic blocks of the final summary.}
    \label{fig:example}
\end{figure*}
Most unsupervised extractive methods are graph-based \cite{zheng-lapata-2019-sentence,dong-etal-2021-discourse,liang-etal-2021-improving,9664266}.
They represent document sentences as nodes in a graph, where the edge value is the similarity between sentences. Then, they measure the importance of each node via computing the degree centrality \cite{radev-etal-2000-centroid} or running PageRank \cite{pagerank} algorithm.
\citet{liang-etal-2021-improving} pointed out that centrality-based methods always tend to select sentences within the same facet (i.e. aspect, sub-topic) and proposed a facet-aware ranking (FAR) method to tackle this problem. FAR forces a centrality-based model to select summary sentences from different facets by incorporating the relevance between the candidate summary and the document.
However, this method faces two problems when the document is extremely long: 1) As the input length increases, the document will have more noise and insignificant facets. The relevance computation between the candidate summary and the document may cause the facet-aware ranking to be influenced by insignificant facets. 2) The running time of FAR will rise rapidly as the number of extracted sentences increases. Due to FAR needs to compute the relevance score number of combinations $C_m^k$ times, where $k$ is the number of extracted summary sentences and $m$ is the number of candidate salient sentences.

To tackle these problems, in this paper, we propose a novel Coarse-to-Fine Facet-Aware Ranking (C2F-FAR) Framework based on semantic blocks, which consists of two stages with different granularities: semantic blocks and sentences. The semantic block means continuing sentences that describe the same facet.
We use a simple example in Fig.~\ref{fig:example} to describe the motivation for building two stages.
Fig.~\ref{fig:example} shows four facet-aware semantic blocks. Each block contains continuous sentences describing the same facet, which is listed on the right. From the coarse-level view, we should first filter blocks with unimportant facets in the document, e.g. the block related to ``additional goal of PILT'' in Fig.~\ref{fig:example}.
Then, from the fine-level view, we should select proper sentences in each block, which are more relevant to the block facet. Note that we only show the most relevant sentences with the facet of each semantic block and omit unrelated sentences due to the space limitation.
Finally, the highlighted sentences should be selected as the summary.

Following this intuitive process, we designed our framework with a coarse-level stage and a fine-level stage. The coarse-level stage aims to select several salient facet-aware semantic blocks for the fine-level stage. We first segment the document into facet-aware semantic blocks by our proposed new document segmentation algorithm, which is inspired by TextTiling \cite{hearst-1997-text}. Then, we filter insignificant facets via a coarse-level centrality estimator to measure the salience of blocks.
The fine-level stage aims to select final summary sentences from previously selected blocks.
We first select candidate sentences in each block to represent its facet by simply computing relevance between sentences and the block.
Finally, we extract the final summary from candidate sentences by sentence-level centrality-based estimator.
Overall, the coarse-level stage can identify all facets of the document effectively and filter insignificant ones. The fine-level stage can reduce the influence of facets with many sentences by only selecting several related sentences for the final ranking. This framework with a hierarchical coarse-to-fine structure can guarantee effective and efficient long document summarization.

We evaluate the effectiveness and efficiency of our C2F-FAR on four long-document summarization datasets with two different metrics. Our method achieves new state-of-the-art performance on Gov-Report and BillSum. It is comparable to strong baselines on arXiv and PubMed. Besides, our method can achieve a speedup of 4-28 times more than two strong baselines.

\section{Methodology} \label{sec:method}
\begin{figure*}[ht]
    \centering
    \includegraphics[width=0.8\textwidth]{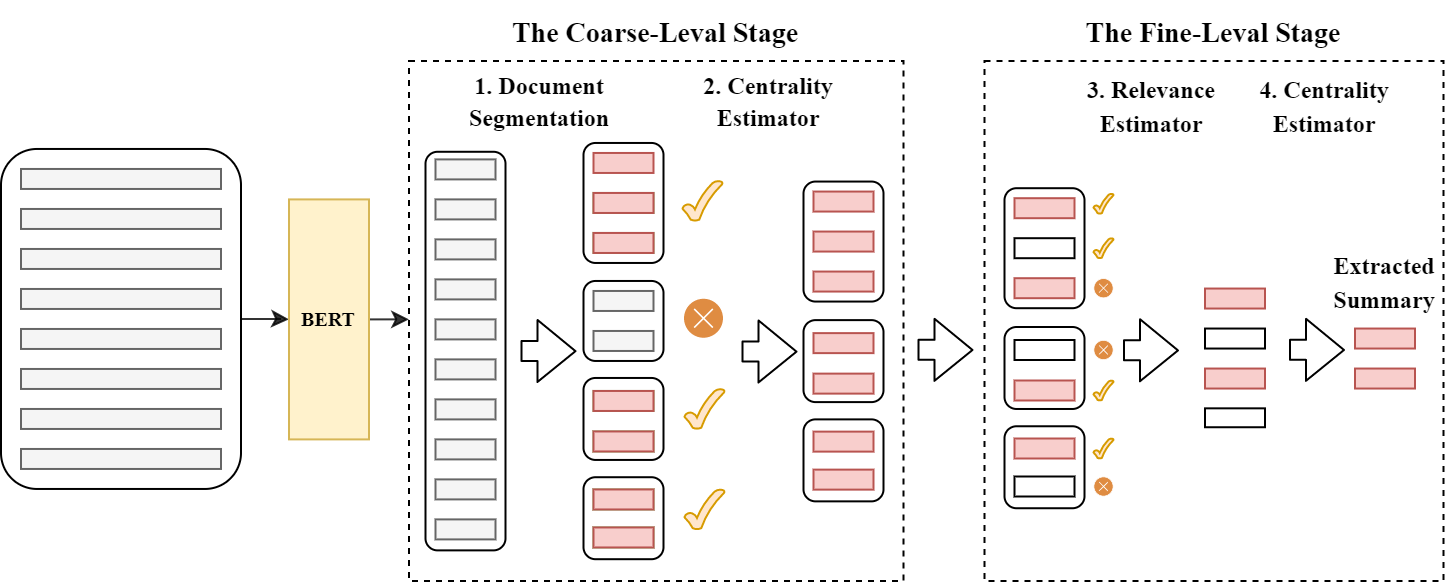}
    \caption{The workflow of our proposed coarse-to-fine facet-aware ranking framework.}
    \label{fig:framework}
\end{figure*}
We show the workflow of our proposed coarse-to-fine facet-aware ranking (C2F-FAR) framework in Fig.~\ref{fig:framework}. After encoding the document into sentence embeddings, the workflow contains two main stages and each stage contains two steps.

(1) In the coarse-level stage, we first employ a document segmentation algorithm to split the document into coarse-level semantic blocks and we call them facet-aware semantic blocks. Then, we score all blocks via the centrality estimator and select top-ranked blocks for the next fine-level stage.

(2) In the fine-level stage, we first select several sentences of each facet-aware semantic block, which can cover the main facet of each block. Then, we employ a sentence-level centrality estimator to score selected sentences and extract the final summary.

We describe the details of each step in the following sections.

\subsection{Sentence Embeddings}
Formally, let $\mathcal D$ indicate a long document containing $n$ sentences $\{s_1,\dots, s_n\}$. In this paper, we employ pre-trained language model to obtain the sentence embeddings $\{v_1,\dots,v_n\}$. Specifically, we employ an improved BERT \cite{devlin2019bert} from previous work PacSum \cite{zheng-lapata-2019-sentence} to represent each sentence $s_i$ with the hidden state $v_i$ of ``[CLS]'' token. This improved BERT can obtain better sentence semantic representation.

\subsection{The Coarse-Level Stage}
The coarse-level stage contains two steps: document segmentation and coarse-level centrality estimator. The document segmentation splits the document into semantic blocks. The coarse-level centrality estimator employs a directed centrality score to measure the importance of each facet-aware semantic block. After the coarse-level stage, we only keep top-ranked $\alpha \times m$ semantic blocks of the whole document, where $m$ is the number of facet-aware semantic blocks and $\alpha$ is a hyper-parameter used to control the ratio of reserved important blocks (default $\alpha=0.5$).

\subsubsection{Document Segmentation Algorithm}
\begin{figure}
\centering
\begin{minipage}[t]{0.48\textwidth}
\centering
\includegraphics[width=5cm]{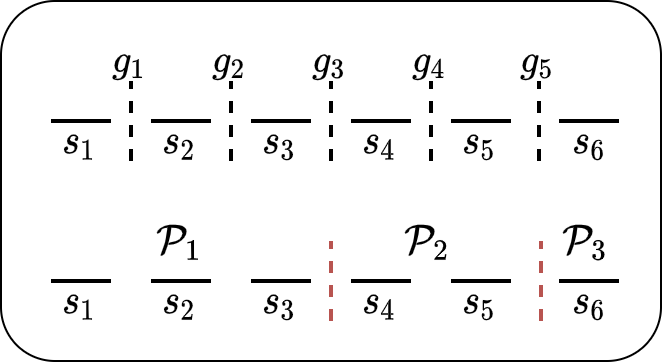}
\caption{A diagram for document segmentation. }
\label{fig:segmentation}
\end{minipage}
\begin{minipage}[t]{0.48\textwidth}
\centering
\includegraphics[width=5.5cm]{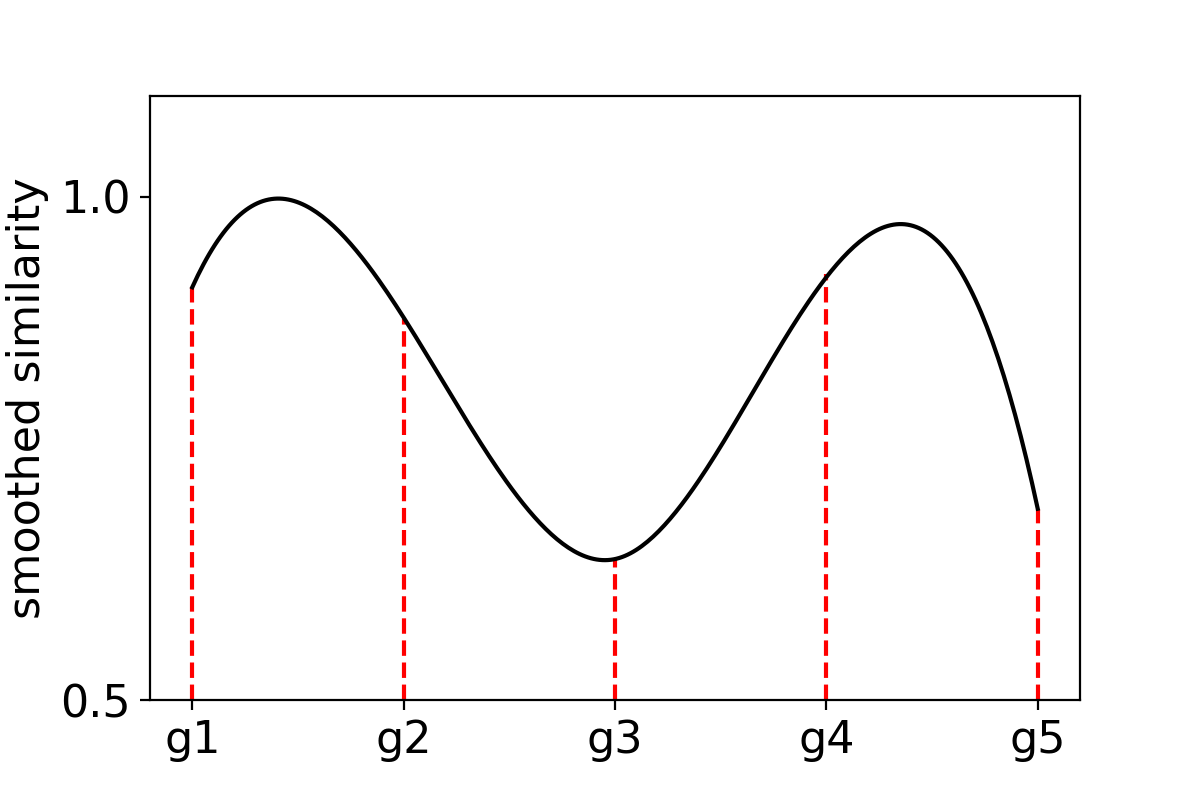}
\caption{The smooth similarity curve.}
\label{fig:sim}
\end{minipage}
\end{figure}
We propose a simple but effective document segmentation algorithm to split the input document into facet-aware semantic blocks. This algorithm is based on the assumption that when sentences with adjacent positions are semantically similar, they focus on the same facet \cite{skorochodko1971}.
As shown in Fig.~\ref{fig:segmentation}, the algorithm aims to select some potential segmentation points to segment the document into several facet-aware semantic blocks $\mathcal P_1=\{s_{p_1^s}, ...,s_{p_1^e}\}$, ..., $\mathcal P_m=\{s_{p_m^s}, ...,s_{p_m^e}\}$.
Our proposed document segmentation algorithm is inspired by TextTiling \cite{hearst-1997-text}. It contains two steps: similarity measure and segmentation point identification. 

In the similarity measure step, we compute the similarity of sentences on both sides of the potential segmentation point $g_i$. 
Each side select $w$ sentences and apply mean operation method over their vectors to obtain global representations $b_i^{l}=\frac{1}{w}\sum_{j=i-w+1}^{i}v_j$ and $b_i^{r}=\frac{1}{w}\sum_{j=i+1}^{i+w}v_j$, where $b_i^l$ and $b_i^r$ refer to the left and right side block with $w$ sentences, respectively. 
The similarity of the sentence on both sides of the potential segmentation point $g_i$ is computed by cosine similarity $sim_i = \frac{b_i^{l}\cdot b_i^{r}}{||b_i^{l}|| ||b_i^{r}||}$.

Then, we apply the moving average on the similarity list of potential segmentation points $\{sim_1, ..., sim_{n-1}\}$ to get a smooth similarity list with Equ. (\ref{eq:1})
\begin{equation}
    \hat{sim}_i=\frac{1}{2\hat{w}+1}\sum_{j=i-\hat{w}}^{i+\hat{w}}sim_j
    \label{eq:1}
\end{equation}
where the $\hat{w}$ is the window size used for moving average operation and the similarity list is refactored as $\{\hat{sim}_1, ..., \hat{sim}_{n-1}\}$. In this paper, the window size $w$ and $\hat{w}$ are all set as 2.

The segmentation point identification step is based on the smooth similarity list. We show an intuitive similarity curve in Fig.~\ref{fig:sim}. If the value of $\hat{sim}_i$ is low, the facets in the left and right blocks are different. So we should segment them with the point $g_i$. We can see that segmentation points $g_3$ and $g_5$ are the local minimum value of the curve in Fig.~\ref{fig:sim}, which are suitable to segment the document.

We convert the similarity list of the potential segmentation point into depth score series $\{d_i\}_{i=1}^{n-1}$ by Equ. \ref{eq:de} to select proper segmentation points.
\begin{equation}
    \begin{split}
        d_i &= \max\{(\hat{sim}_{i-1} - \hat{sim}_{i}), 0\}\\
        &+ \max\{(\hat{sim}_{i+1} - \hat{sim}_{i}), 0\}
    \end{split}
    \label{eq:de}
\end{equation}
When the similarity of the potential segmentation point is the local minimum value, it will become the local maximum value after being converted into a depth score.
If $d_i > \epsilon$, we choose the potential segmentation point $g_i$ as the segmentation point.
The $\epsilon$ is a threshold and is decided by the mean $\mu$ and standard deviation $\sigma$ of the depth score series. We set $\epsilon=\mu + \lambda\cdot\sigma$, where $\lambda$ is a hyper-parameter to control the granularity of segmentation. The greater the $\lambda$, the segmented block contains more sentences.

Finally, we can segment the whole document into some facet-aware semantic blocks $\mathcal P_1=\{s_{p_1^s}, ...,s_{p_1^e}\}$, ..., $\mathcal P_m=\{s_{p_m^s}, ...,s_{p_m^e}\}$, like examples in the Fig.~\ref{fig:segmentation}.

\subsubsection{Coarse-Level Centrality Estimator}
We introduce the coarse-level centrality estimator for filtering unimportant facet-aware semantic blocks in this section. 
We represent the semantic information of each block $\mathcal P_i$ by computing the average of sentence vectors contained in the block.
\begin{equation}
    p_i = \frac{1}{|\mathcal P_i|}\sum_{s_i \in \mathcal P_i}(s_i)
\end{equation}
The representations of blocks are $\{p_1,\dots,p_m\}$. Then, we employ directed centrality \cite{zheng-lapata-2019-sentence} to score each block based on the assumption that the contribution of any two nodes’ connection to their respective centrality is influenced by their relative position.
\begin{equation}
    \mathcal{C}(p_i) = \lambda_1\sum_{j<i}^n{p_i\cdot p_j} + \lambda_2\sum_{j>i}^n{p_i\cdot p_j}
\end{equation}
After that, we rank all blocks via directed centrality score $\mathcal{C}(p_i)$ and only keep top-ranked $\alpha$ percent semantic blocks for the next fine-level stage, where $\alpha$ is a hyper-parameter to control the ratio of reserved blocks.

\subsection{The Fine-Level Stage}
The fine-level stage contains two steps: relevance estimator and fine-level centrality estimator. The relevance estimator is used to select some sentences in each facet-aware semantic block, which can retain the main information of the block. The fine-level centrality estimator is applied to sentences from the previous relevance estimator and also employs the directed centrality score to extract the final summary.

\subsubsection{Relevance Estimator}
The relevance estimator simply computes the relevance between sentences and the block to select sentences to represent the facet in semantic blocks. This step is based on the assumption that each facet-aware semantic block only contains one facet. 
We employ cosine similarity to measure the relevance between sentence representation $v_j$ and block representation $p_i$.
\begin{equation}
    \mathcal R(s_j) = \frac{v_j\cdot p_i}{||v_j|| ||p_i||}, s_j\in \mathcal P_i
\end{equation}
For each semantic block, we select top-ranked $\beta$ sentences, where $\beta$ is the average number of semantic block sentences, which is determined by the granularity of document segmentation. If the number of sentences in a block is lower than $\beta$, we keep all sentences. Then, we can get $t$ candidate sentences $\{\hat{s}_1,\dots,\hat{s}_t\}$ for the final summary selection.

\subsubsection{Fine-Level Centrality Estimator}
The final fine-level centrality estimator aims to select the final summary sentences from previous candidate sentences.
The final fine-level centrality estimator measures the importance of each candidate sentence as follows:

\begin{equation}
            \mathcal{C}(s_i) = \lambda_1\sum_{j<i}^t{v_i\cdot v_j} + \lambda_2\sum_{j>i}^t{v_i\cdot v_j} 
\end{equation}
where $s_i,s_j \in \{\hat{s}_1,\dots,\hat{s}_t\}$.
We select top-ranked $k$ sentences as the final summary, where $k$ is the average number of sentences of different datasets.

\section{Experiments}

\subsection{Datasets}
\begin{table}[ht]
\centering
\small
\begin{tabular}{l|c|cc|cc}
\hline
\multirow{2}{*}{Datasets} & \multirow{2}{*}{\#docs} & \multicolumn{2}{c|}{document} & \multicolumn{2}{c}{ summary} \\
           &       & words & sen. & words & sen. \\ \hline
Gov-Report & 973   & 9,409 & 304  & 657   & 23   \\
BillSum    & 3,269 & 2,148 & 169  & 209   & 10    \\
arXiv      & 6,440 & 4,938 & 206  & 220   & 10   \\
PubMed     & 6,658 & 3,016 & 107  & 203   & 8    \\ \hline
\end{tabular}
\caption{Statistics information of Gov-Report, BillSum, arXiv, and PubMed datasets. We compute the average document and summary length in terms of words and sentences, respectively.}
\label{tab:datasets}
\end{table}

We evaluate our C2F-FAR on 4 datasets. The statistics information of them is shown in Tab. \ref{tab:datasets}.

\textbf{Gov-Report} \cite{huang-etal-2021-efficient} is a large-scale long document summarization dataset containing 19,466 long reports published by U.S. Government Accountability Office (GAO) and Congressional Research Service (CRS). Documents and summaries in Gov-Report are significantly longer than other datasets.

\textbf{BillSum} \cite{kornilova-eidelman-2019-billsum} contains US Congressional bills and human-written references from the 103rd-115th (1993-2018) sessions of Congress. We found that previous works have some errors in the sentence segmentation of the dataset. We re-segmented this dataset with the StanfordNLP toolkit and conducted experiments on the basis of the new sentence segmentation. 

\textbf{arXiv} and \textbf{PubMed} \cite{cohan-etal-2018-discourse} are two long scientific document summarization datasets from scientific papers.

\subsection{Settings and Metrics}
We employ sentence-BERT\footnote{https://github.com/huggingface/transformers} from \cite{zheng-lapata-2019-sentence} to encode sentences in the document, which converts each sentence into a vector with 768 elements. The window size of the document segmentation algorithm is 2.
The default setting of $\lambda$ is 1.0 and $\alpha$ is 0.5.

We reported ROUGE-1/2/L scores with \texttt{ROUGE-1.5.5.pl} script\footnote{https://github.com/andersjo/pyrouge} \cite{lin-2004-rouge} and BertScore \cite{bert-score} of baselines and our methods. The ROUGE score is the lexical level metric to measure the similarity between extracted summary and gold summary. The BertScore\footnote{https://github.com/Tiiiger/bert\_score} measures the semantic level similarity between the extracted summary and gold reference. 

\subsection{Baselines}
We compare our method with recent strong unsupervised extractive summarization models. 

\textbf{Lead}, which selects the first $k$ tokens as a summary.

\textbf{Oracle}, which is the upper bound of extractive summarization methods. It selects sentences by computing ROUGE scores with the gold summary.

\textbf{TextRank} \cite{mihalcea-tarau-2004-textrank} and \textbf{LexRank} \cite{erkan-2004-lexrank}, which are two traditional unsupervised ranking method based on TF-IDF and PageRank algorithm to select salient sentences.

\textbf{TextRank(BERT)}, which employs embeddings from improved BERT to compute the edge weight of TextRank.

\textbf{FAR} \cite{liang-etal-2021-improving}, which defined the facet bias problem and proposed a facet-aware centrality method to tackle the bias problem. 

\subsection{Evaluation of Summary Quality and Inference Time}

We report the results of automatic and human evaluation of all systems to measure the extracted summary quality of our C2F-FAR. Besides, we also compare the inference time of our method with two strong baselines to prove the high efficiency of our method.
\begin{table*}[ht]
\centering
\small
\begin{tabular}{l|cccc|cccc}
\hline
\multicolumn{1}{c|}{\multirow{2}{*}{Models}} &
  \multicolumn{4}{c|}{Gov-Report} &
  \multicolumn{4}{c}{BillSum} \\
\multicolumn{1}{c|}{} & R-1   & R-2   & R-3   & BS-F  & R-1   & R-2   & R-3   & BS-F  \\ \hline
Oracle                & 74.87 & 49.02 & 72.48 & 88.83 & 65.24 & 47.09 & 58.81 & 86.29 \\
Lead                  & 50.94 & 19.53 & 48.45 & 83.47 & 40.53 & 18.28 & 34.15 & 80.24 \\
LexRank               & 40.16 & 8.85  & 37.65 & 82.48 & 34.39 & 10.05 & 28.93 & 79.76 \\
TextRank(TF-IDF)      & 53.19 & 23.12 & 49.86 & 84.83 & 40.04 & 16.12 & 32.64 & 80.81 \\
TextRank(BERT)        & 56.00 & 22.42 & 52.86 & 85.10 & 38.05 & 12.99 & 31.46 & 80.02 \\ \hline
PacSum                & 56.89 & 26.88 & 54.33 & 85.02 & 41.11 & 17.24 & 34.54 & 81.33 \\
FAR                   & 57.51 & 27.54 & 54.94 & 85.38 & 41.53 & 17.44 & 34.84 & 81.21 \\
C2F-FAR &
  \textbf{57.98} &
  \textbf{27.63} &
  \textbf{55.33} &
  \textbf{86.62} &
  \textbf{42.53} &
  \textbf{17.85} &
  \textbf{35.58} &
  \textbf{81.57} \\ 
\hline
\end{tabular}
\caption{Results on Gov-Report and BillSum test set. BS-F refers to $F_1$ of the BertScore.}
\label{tab:main_res_1}
\end{table*}

\begin{table*}[]
\centering
\small
\begin{tabular}{l|cccc|cccc}
\hline
\multicolumn{1}{c|}{\multirow{2}{*}{Models}} & \multicolumn{4}{c|}{arXiv}    & \multicolumn{4}{c}{PubMed}    \\
\multicolumn{1}{c|}{} &
  \multicolumn{1}{c}{R-1} &
  \multicolumn{1}{c}{R-2} &
  \multicolumn{1}{c}{R-3} &
  \multicolumn{1}{c|}{BS-F} &
  \multicolumn{1}{c}{R-1} &
  \multicolumn{1}{c}{R-2} &
  \multicolumn{1}{c}{R-3} &
  \multicolumn{1}{c}{BS-F} \\ \hline
Oracle                & 53.88 & 23.05 & 34.9  & 87.06 & 55.05 & 27.48 & 38.66 & 87.05 \\
Lead                  & 33.66 & 8.94  & 22.19 & 82.97 & 35.63 & 12.28 & 25.17 & 80.43 \\
LexRank               & 33.85 & 10.73 & 28.99 & 80.42 & 39.19 & 15.87 & 34.53 & 83.21 \\
TextRank(TF-IDF)      & 36.59 & 10.06 & 30.29 & 82.49 & 38.66 & 15.87 & 34.53 & 82.43 \\
TextRank(BERT)        & 34.68 & 8.78  & 30.05 & 81.19 & 39.43 & 12.89 & 34.66 & 83.39 \\ \hline
PacSum                & 38.58 & 11.12 & 33.5  & 81.78 & 39.79 & 14.00    & 36.09 & 83.43 \\
FAR &
  \textbf{40.92} &
  \textbf{13.75} &
  \textbf{35.56} &
  \textbf{83.74} &
  \textbf{41.98} &
  \textbf{16.74} &
  \textbf{37.58} &
  \textbf{83.89} \\
C2F-FAR               & 39.32 & 11.65 & 34.28 & 82.04 & 40.12 & 14.79 & 36.91 & 83.50 \\ \hline
\end{tabular}
\caption{Results on arXiv and PubMed test set. BS-F refers to $F_1$ of the BertScore.}
\label{tab:main_res_2}
\end{table*}

The automatic evaluation results of ROUGE score and BertScore are shown in the Tab. \ref{tab:main_res_1} and Tab. \ref{tab:main_res_2}. These two scores measure the lexical and semantic level similarity between extracted summary and gold reference, respectively.
All reported results of our C2F-FAR framework employed the default hyper-parameters $\lambda=1$ and $\alpha=0.5$.
We can see that our C2F-FAR achieved new state-of-the-art results on Gov-Report and BillSum in unsupervised methods. 
The performance of our method also is better than PacSum and comparable to FAR on the other two datasets: arXiv and PubMed. 
We will analyze the reason for the results on arXiv and PubMed in the discussion section.
Interestingly, there is no big difference between the two versions of TextRank. We guess that the iterative algorithm based on PageRank is not sensitive to the similarity measure methods.

To evaluate the ability of our C2F-FAR in reducing facet bias and improving the quality of extracted summaries, we asked 3 human annotators to evaluate the extracted summaries of C2F-FAR and FAR with the gold reference summary.
Three annotators were given extracted and gold summary. Then they were asked to give 0-2 scores for facets coverage (whether the extracted summary contains most primary facets) and quality (the comprehensive feelings of the extracted summary) of 20 random sampled examples from test sets of BillSum and 20 random sampled examples from test sets of Gov-Report (0-bad, 1-normal, 2-good).
The results of FAR in terms of facets coverage is 1.16 and quality is 1.03.
Our C2F-FAR performs significantly better (p < 0.05 with Mann-Whitney U tests) than FAR whose facets coverage is \textbf{1.38} and quality is \textbf{1.15}. 

\begin{figure}[ht]
    \centering
    \includegraphics[width=0.45\textwidth]{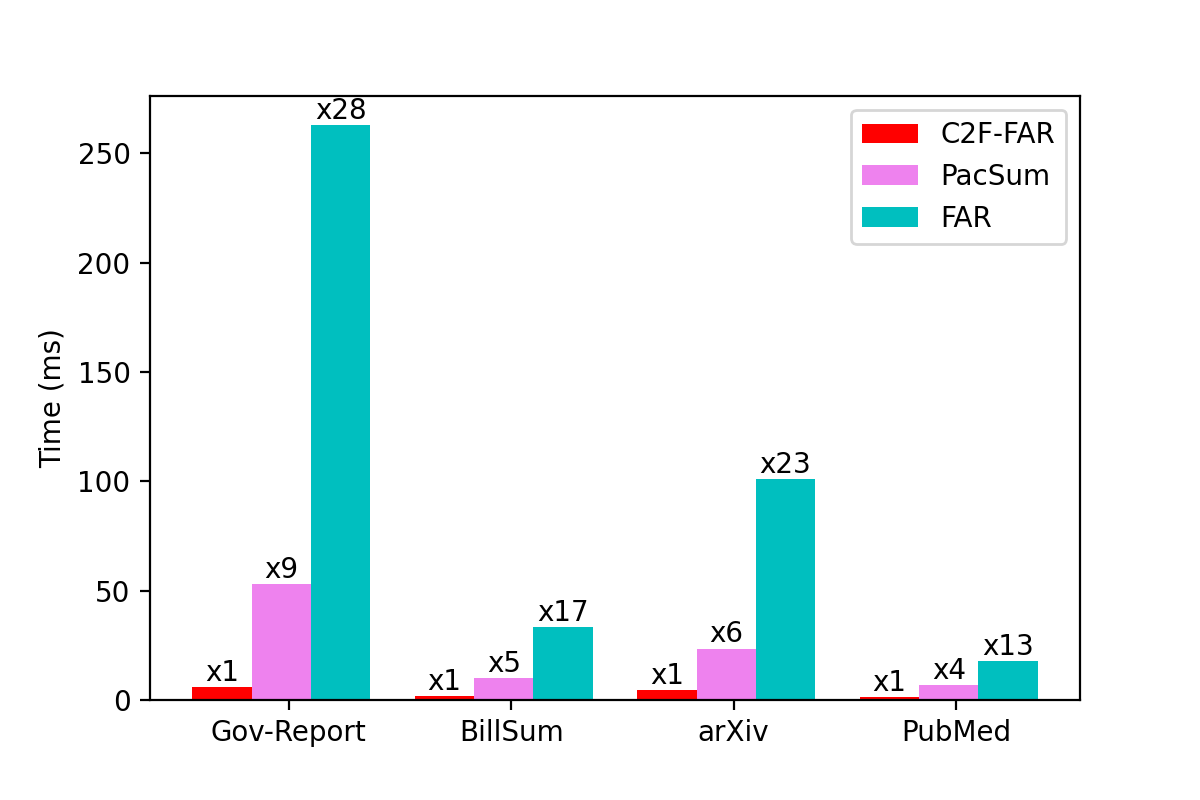}
    \caption{The inference time of each system. Each time is the average of multiple runs (10 times). ''$\times N$`` means the running time is $N$ times (rounded up) of our method.}
    \label{fig:time}
\end{figure}

To test the inference time of our method, we randomly select 100 examples from the test set of each dataset and ensure that the average input length of these 100 examples is the same as the average length of the test set.
Then, we run each method 10 times and report the average inference time of them on four datasets.
We can see Fig.~\ref{fig:time} and find that our method is far ahead of the other two methods in inference time, and this advantage becomes more obvious as the length of the input document increases.


Overall, compared with other methods, our method takes into account both efficiency and effectiveness.
In addition, our framework also can adjust the specific ranking method in each step for datasets with different types and domains, which makes it flexible.

\section{Analysis}
In this section, we first analyze the parameter sensitivity of our C2F-FAR and then discuss the reason why our method is inferior to the FAR on arXiv and PubMed via facets analysis of extracted sentences. 

\subsection{Impact of Hyper-parameters}

\begin{table}[ht]
\centering
\small
\begin{tabular}{c|ccc|ccc}
\hline
Datasets  & \multicolumn{3}{c|}{BillSum} & \multicolumn{3}{c}{Gov-Report} \\ \hline
$\lambda$ & $\beta$   & Para.  & Comp.  & $\beta$   & Para.   & Comp.   \\ \hline
0         & 3         & 70      & 41\%    & 3         & 120      & 39\%     \\
0.5       & 4         & 45      & 27\%    & 5         & 74       & 24\%     \\
1         & 6         & 27      & 16\%    & 10        & 44       & 14\%     \\
1.5       & 11        & 15      & 9\%     & 15        & 26       & 9\%      \\
2         & 20        & 8       & 5\%     & 20        & 15       & 5\%      \\
2.5       & 36        & 4       & 2\%     & 36        & 4        & 1\%      \\ \hline
\end{tabular}
\caption{Parameters affected by $\lambda$ on two datasets. Para. means the average number of blocks with different hyper-parameters $\lambda$. Comp. means the ratio of the number of blocks to the number of sentences. $\beta$ is the average number of sentences in a block.}
\label{tab:params}
\end{table}

In this section, we will analyze the parameter sensitivity of two hyper-parameters in our C2F-FAR framework: 1) $\lambda$ is used to control the granularity of the document segmentation algorithm; 2) $\alpha$ is used to control the ratio of reserved blocks of the coarse-level centrality estimator. We can see the relationship between compression ratio and $\lambda$ in the Tab. \ref{tab:params}. The default setting of $\lambda=1$ has an impressive compression ratio on two datasets.

\begin{figure}[ht]
    \centering
    \includegraphics[scale=0.45]{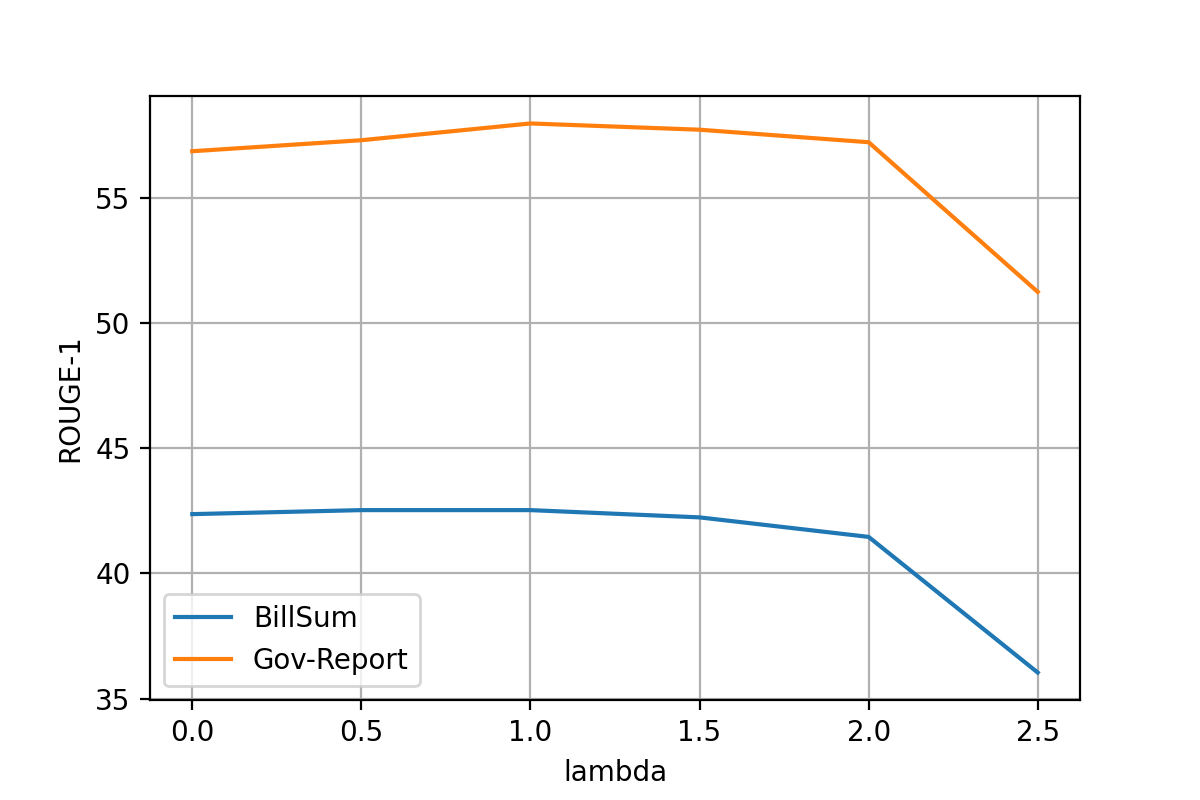}
    \includegraphics[scale=0.45]{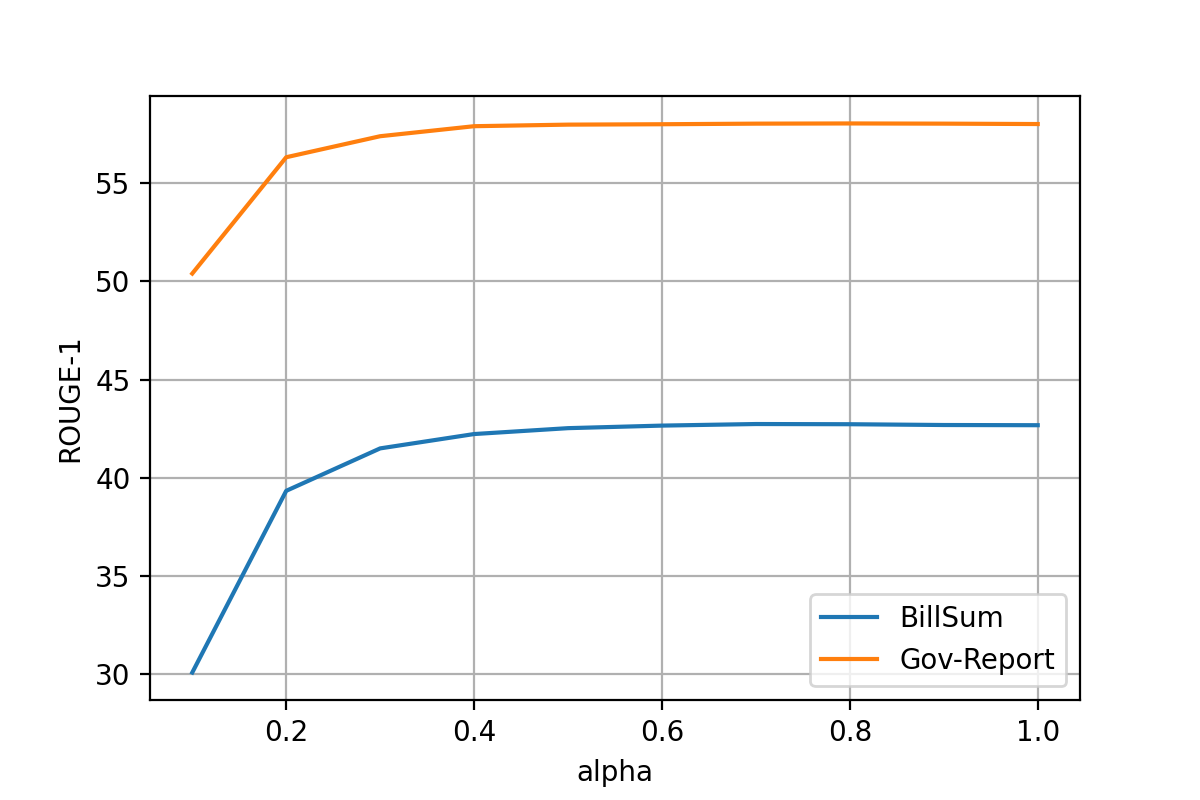}
    \caption{Impact of hyper-parameters $\lambda$ and $\alpha$.}
    \label{fig:hyps}
\end{figure}

We fix $\alpha$ and show the change of the ROUGE-1 score while $\lambda$ changes in Fig.~\ref{fig:hyps}. We can find that the performance is best when $\lambda=1.0$, and there is little change when $\lambda \in [0, 2]$.
This shows that our algorithm is stable. You can set a larger $\lambda$ to get a faster running speed while ensuring good performance. 
We set the value range of $\lambda$ between 0.0 and 2.5 because when $\lambda$ is less than 0, the most segmented blocks contain one sentence. Then the following algorithms are equivalent to acting on the sentence-level structure.

We also fix $\lambda$ and show the change of the ROUGE-1 score while $\alpha$ changes in Fig.~\ref{fig:hyps}. We can see that the second half of the curve is almost flat. 
This shows that the low centrality score of the segmented segment does not contribute to the final summary quality. The facets contained in these blocks are not important to the whole document. We can filter them with $\alpha$ in the coarse-level step and achieve a faster running speed.

The analysis of the two hyper-parameters proves that our C2F-FAR framework can employ simple hyper-parameter settings to improve the running speed of the algorithm while ensuring the quality of the summary.

\subsection{Facets of Extracted Sentences}
\begin{table}[ht]
\centering
\small
\begin{tabular}{c|cccc}
\hline
               & Gov-Report & BillSum & arXiv & PubMed \\ \hline
\#fac.        & 11.1       & 7.0     & 3.8   & 3.2    \\
\#sen.        & 20         & 10      & 10    & 7      \\
\#sen./\#fac. & 1.80       & 1.42    & 2.63  & 2.19   \\ \hline
\end{tabular}
\caption{\#fac. refers to the average number of facet-aware semantic blocks, which contain extracted sentences. \#sen. refers to the number of extracted sentences. \#sen./\#fac. refers to the average number of sentences from each block. Extracted sentences are from the Oracle system.}
\label{tab:bad_case}
\end{table}
In Tab. \ref{tab:bad_case}, we employ the extracted sentences from the Oracle system to analyze the characteristics of four datasets. The granularity of the document segmentation algorithm is $\lambda=1$. 
We can see that selected summary sentences in arXiv and PubMed datasets distribute in fewer facet-aware semantic blocks than those in Gov-Report and BillSum. Our model tends to select summary sentences from more blocks, thus achieving better performance in Gov-Report and BillSum datasets.

By observing the extracted summary sentences from the Oracle system and combining the results in Tab. \ref{tab:bad_case}, we can roughly get the reason why our model is not as good as FAR on these datasets: the contents of the document and the summary is more concentrated on 3-4 facets of the document. 
Besides, the extracted sentences of them are mainly distribute at the start or end part (introduction and conclusion) of the document \cite{dong-etal-2021-discourse}. 
However, our method is more inclined to select summary sentences from more blocks and select many sentences in the middle part of the document. This leads to our method not performing so well on these two datasets.

\section{Related Work}
\subsection{Long Document Summarization}
Thanks to the development of Transformer-based \cite{vaswani-2017-attention} Pre-trained Language Models (PLMs), such as BERT \cite{devlin-etal-2019-bert}, recent summarization models \cite{liu-lapata-2019-text,zhang2019pegasus,li-etal-2020-leveraging-graph,lewis-etal-2020-bart,zhong-etal-2020-extractive,liu-liu-2021-simcls,liu-etal-2021-refsum} achieved excellent performance in short document summarization. 
However, these models can not be simply transferred to long document summarization due to both salient and noise content increasing according to the increase of the input text. 
How to summarize the long-form document, including books \cite{mihalcea-ceylan-2007-explorations}, patents \cite{sharma-etal-2019-bigpatent}, scientific publications \cite{qazvinian-radev-2008-scientific,cohan-etal-2018-discourse}, etc., is an important and long-standing challenge.

Most recent works for long-form document summarization are supervised and mainly tackle this problem through two angles. The first angle tends to design more efficient self-attention mechanisms to reduce the complexity. \cite{child2019generating,kitaev2020reformer,beltagy2020longformer,zaheer2020bigbird,huang-etal-2021-efficient,tay2021long} 
The other angle employed the condense-then-generate paradigm \cite{cohan-etal-2018-discourse,xu-durrett-2019-neural,zhang-etal-2019-hibert,lebanoff-etal-2019-scoring,zhu-etal-2020-hierarchical,akiyama-etal-2021-hie,grail-etal-2021-globalizing}. This paradigm first employs sentence/discourse-level structure to select salient sentences and then generates the summary based on them. 
This paradigm is intuitive and similar to the behavior of humans summarizing a long document. Our method also borrows some ideas from it.

\subsection{Unsupervised Summarization}
Most traditional unsupervised summarization methods are graph-based and extractive \cite{radev-etal-2000-centroid,mihalcea-tarau-2004-textrank,radev-etal-2000-centroid,erkan-2004-lexrank,wan-2008-exploration}. They represent the document as a graph, where each sentence is a node with a weighted edge which is the similarity between nodes. They rank sentences via computing centrality with node degree or PageRank algorithm \cite{pagerank}. Recently, many unsupervised works \cite{pmlr-v97-chu19b,zhou-rush-2019-simple,zheng-lapata-2019-sentence,yang-etal-2020-ted,xu-etal-2020-unsupervised,liu-2021-unsupervised,dong-etal-2021-discourse,liang-etal-2021-improving} combined traditional methods with PLMs and achieved fantastic performance.

\newcite{zheng-lapata-2019-sentence} first employed BERT to enhance similarity measure for graph-based ranking and proposed a directed degree centrality computation method. \newcite{dong-etal-2021-discourse} pointed out that the previous method is not suitable for long scientific papers and proposed a hierarchical discourse-based unsupervised ranking method. \newcite{liang-etal-2021-improving} found that they all ignored the facet-bias problem \cite{mao-etal-2020-facet}, which is ubiquitous in unsupervised methods and proposed a facet-aware ranking method FAR. However, as the document length increases, they cannot extract proper sentences which cover vital facets of the document, from rapidly increased insignificant facets.

\section{Conclusion}
In this paper, we focus on unsupervised long document summarization tasks, which is a vital and long-standing challenge in text summarization. To obtain summary sentences efficiently and effectively, we proposed a novel coarse-to-fine facet-aware ranking framework. Our method can achieve new state-of-the-art results on two datasets. Experiments show that our approach is effective and efficient for the long document summarization task. In future work, we will investigate how to refactor this process into an end-to-end paradigm.

\section*{Acknowledgements}
This work was supported in part by the National Natural Science Foundation of China (Grant Nos.U1636211, 61672081,61370126), the 2020 Tencent Wechat Rhino-Bird Focused Research Program, and the Fund of the State Key Laboratory of Software Development Environment (Grant No. SKLSDE-2021ZX-18).

\bibliography{anthology,custom}
\bibliographystyle{acl_natbib}




\end{document}